\documentclass[conference]{IEEEtran}

\usepackage{cite}
\usepackage{amsmath,amssymb,amsfonts}
\usepackage{algorithm}
\usepackage{algpseudocode}
\usepackage{textcomp}
\usepackage{xcolor}

\providecommand{\doi}[1]{doi: {\footnotesize \href{http://dx.doi.org/#1}{\path{#1}}}}

\usepackage[pdftex=true,breaklinks=true,hidelinks=true,colorlinks=true,citecolor=blue]{hyperref}

\usepackage{graphicx}
\graphicspath{{pdf/}{photo/}{figures/}}
\DeclareGraphicsExtensions{.eps,.jpeg,.jpg,.png,.pdf}

\usepackage{subcaption}
\usepackage[]{caption}

\usepackage{fancyhdr}
\usepackage{booktabs}

\def\BibTeX{{\rm B\kern-.05em{\sc i\kern-.025em b}\kern-.08em
    T\kern-.1667em\lower.7ex\hbox{E}\kern-.125emX}}
\begin{document}

\makeatletter
\fancypagestyle{plain}{
    \fancyhf{}
    \fancyhead[C]{\emph{Preprint submitted to 2\textsuperscript{nd} Int. Sci. Conf. ALFATECH – Smart Cities and Modern Technologies 2025}}
    \fancyfoot[C]{\thepage}
     \renewcommand{\headrulewidth}{0pt}
}
\makeatother

\makeatletter
\fancypagestyle{custom}{
    \fancyhf{}
    \fancyfoot[C]{\thepage}
    \renewcommand{\headrulewidth}{0pt}
}
\makeatother

\title{Person detection and re-identification in open-world settings of
retail stores and public spaces\\
\thanks{Identify applicable funding agency here. If none, delete this.}
}

\author{\IEEEauthorblockN{Branko Brkljač}
\IEEEauthorblockA{\textit{Dept. of Power, Electronic and Telecommunication Engineering} \\
\textit{Faculty of Technical Sciences, University of Novi Sad}\\
Novi Sad, Republic of Serbia \\
brkljacb@uns.ac.rs}
\and
\IEEEauthorblockN{Milan Brkljač}
\IEEEauthorblockA{\textit{Faculty of Finance, Banking and Auditing} \\
\textit{Alfa BK Univeristy}\\
Novi Beograd, Republic of Serbia \\
milan.brkljac@alfa.edu.rs}
}

\maketitle
\thispagestyle{plain} 

\begin{abstract}
Practical applications of computer vision in smart cities usually assume system integration and operation in challenging open-world environments. In the case of person re-identification task the main goal is to retrieve information whether the specific person has appeared in another place at a different time instance of the same video, or over multiple camera feeds. This typically assumes collecting raw data from video surveillance cameras in different places and under varying illumination conditions. In the considered open-world setting it also requires detection and localization of the person inside the analyzed video frame before the main re-identification step. With multi-person and multi-camera setups the system complexity becomes higher, requiring sophisticated tracking solutions and re-identification models. In this work we will discuss existing challenges in system design architectures, consider possible solutions based on different computer vision techniques, and describe applications of such systems in retail stores and public spaces for improved marketing analytics. In order to analyse sensitivity of person re-identification task under different open-world environments, a performance of one close to real-time solution will be demonstrated over several video captures and live camera feeds. Finally, based on conducted experiments we will indicate further research directions and possible system improvements.
\end{abstract}


\section{Introduction}
\label{Introduction}
\pagestyle{custom}
Modern computer vision techniques are solving complex technical challenges posed by applications in open-world settings. This refers to scenarios in which the system must overcome uncontrolled and difficult to predict characteristics of the operating environment. This is also the case with person detection \cite{dollar2011pedestrian} and person re-identification \cite{gong2014person} tasks, which have traditionally been researched in the context of autonomous vehicles \cite{yurtsever2020survey}, robotics \cite{kwon2021comparative}, human-machine interfaces \cite{behera2021futuristic}, crowd analysis \cite{chen2015person}, and most commonly security applications \cite{gong2011security}. The concept of smart cities \cite{yang2023cooperative} and the integration of such vision systems \cite{almazan2018re, shili2024you, del2024analyzing} into complex decision-making frameworks \cite{quintana2016improving, liu2018tar, ghose2022real} have introduced the need for efficient and robust solutions, with the ability to make necessary trade-offs between operational costs and system performance on a case-by-case basis. The common goal is providing necessary insights into persons' behaviour based on the ability to detect and possibly re-identify the same person at different points in time and space \cite{martini2020open}. Such applications also raise the questions of ethical and privacy nature, which is a significant line of current research \cite{van2020anonymized, hashemifard2023fallen}. Nevertheless, integration of behaviour analysis systems in retail stores and public spaces is gaining momentum \cite{Pietrini2019} and is expected to have significant impact on insightful business operations and space management.

Approaching potential customers based on research analysis of their behaviour can be considered a key ingredient for designing successful marketing strategies and campaigns. It is also manifestation of the general trend of tailored campaigns and data-driven decisions emerging across different modalities of interaction with retail shoppers and public space users. In other words, persons' movement patterns and time spent in different parts of a store or elsewhere directly answer the questions: "How do consumers shop?", or "How people engage in public space?". Data driven insights into these questions are foundations for comprehensive understanding of their behaviour. Thus, behavioural constructs in different conceptual frameworks, like e.g. emotional regulation consumption or compensatory consumption \cite{lee2015emotional}, can be better related with the motivational and emotional aspects of the shopping process and public infrastructure engagement. It is also the kind of data that cannot be obtained from point-of-sale \cite{yamashiro2023customer} or simple visitor statistics \cite{becattini2022mall}. Therefore, spatial analysis of human behavior can be considered one of the complementary technologies for achieving optimal decisions and reducing the cost of adapting to individual needs of customers or users. In this paper, we provide a brief overview of methods and approaches for addressing design challenges for person detection and person re-identification in open-world environments and describe their use in retail sector and public spaces. We also present a demonstration of one cost efficient, close to real-time solution based on OAK-D~lite \cite{OakDlite2025} embedded vision platform \cite{OakDlite2022} and discuss advantages and limitations of such hardware implementation and baseline algorithmic approach.

The rest of the paper is organized as follows. In Section~\ref{Related work} we break down the landscape of existing methods for person detection and person re-identification tasks and provide a structured overview of research directions that have been established by pursuing different solution architectures under the assumptions of specific imaging modalities, identification scenarios, robustness requirements and model learning strategies. In Section~\ref{Proposed person ReID implementation} we go into implementation details of developed low-cost solution and indicate advantages and limitations of the adopted fast prototyping system design approach.  Section~\ref{Experimental results and discussion} presents conducted experiments and discusses their implications for possible applications in retail marketing analytics and public space management. Finally, in Section~\ref{Conclusion} we indicate directions for future research and possible system improvements. Code implementation is available at: {\url{https://github.com/brkljac/personReID}}.

\section{Related work}
\label{Related work}

Person detection \cite{dollar2011pedestrian, shili2024you, del2024analyzing}, alongside face detection \cite{zafeiriou2015survey, feng2022detect}, is probably one of the most studied object detection tasks in the literature \cite{zou2023object}. Similarly, person re-identification (ReID) \cite{gong2014person}, alongside face recognition \cite{wang2022survey}, is one of the long-standing recognition problems in vision \cite{qian2024identifying, zahra2023person}. Its goal is to retrieve information whether the specific person has already appeared in video or over multiple camera feeds in the past. In such case, re-identification  is considered as successful if we can correctly assign person's identity with some of the identities already present in the gallery, or create a new identity entry in the case of novel person in the scene. Described working scenario corresponds to the open-set ReID task, which is different from the closed-set setting in which it is assumed that the query object already exists in the gallery. These two problems, detection and re-identification, can be approached either traditionally, i.e. separately \cite{yaghoubi2021sss, ming2022deep}, in a sequential manner, or jointly, i.e. in an end-to-end manner \cite{zheng2017person, xiao2017joint}, which is a characteristic of more recent ReID approaches. In the following we provide a brief overview of methods utilized in these two specific tasks of interest.

\subsection{Detection task}
\label{Detection task}

In general, besides the problem of defining object categories \cite{alexe2013objectness}, object detection approaches solve two fundamental problems in vision: localization and classification of object instances inside the scene. Consequently, binary classification of persons in the image can be formulated as a two stage: localization and detection problem, or as a single stage, i.e. one step regression task \cite{zou2023object}.  As there are numerous approaches for person detection, their applicability depends on the cost-performance trade-off put in front of system designer. The role of the detector in the broader context of the system in which it operates is also important. In this sense, the choice of an adequate working point on the detector's  receiver operating characteristic (ROC) is also a matter of compromise and the requirements set regarding the subsequent processing of generated information. Thus, precision and recall, or the relationship between detector's sensitivity (hit rate) and specificity (false positives rate) play a crucial role in the design choice of accompanying or downstream processing elements that are relying on detector's decisions. However, in some cases, greedy approaches in which the possibility of a miss is avoided by increasing the sensitivity of the detector, despite a higher number of false alarms, are also justified if it is possible to adequately use such results. Such techniques may include the use of detection or confidence scores in the case of still images, spatial filtering based on the temporal history of the object's locations in the case of video signal, or additional side information obtained from other sources. In the given context of public spaces monitoring and retail analytics, it is interesting to note that usual detection paradigm can be modified in order to overcome some of the difficulties posed by the application scenario. Thus, a new object detection paradigm called Locount is proposed in \cite{cai2021rethinking}, with the goal of localizing groups of objects of interest and estimating number of instances within the group. It is similar to the type of problems considered in \cite{chen2015person}, which indicates that the task of person detection is still of active research interest. This is especially true in the case of open-world environments, where adaptation and resilience to various aggravating circumstances such as illumination variation, image noise, presence of occlusions, and variations in scale or resolution of persons in the video are often needed. It is achieved through advanced learning techniques, data augmentation, or combining of different signal modalities, \cite{li2023indoor, luna2021people}.

\subsection{Re-identification  task}
\label{Re-identification  task}

According to \cite{yadav2024deep}, image-based person re-identification methods can be broadly grouped by: 1)~signal modality (e.g. visible, infrared \cite{lin2022learning}, thermal \cite{li2022confidence}, or cross-modal \cite{huang2023deep}); 2)~learning approaches (e.g. whether the recognition is based on metric learning \cite{zou2021person, wojke2018deep} or more complex deep ReID architectures and learning frameworks \cite{ye2021deep, ming2022deep}); and 3)~how generalization across multiple domains, i.e. cross-domain ReID is addressed \cite{deng2023harmonious}.  Video-based person ReID methods \cite{liu2021video, zang2022multidirection, wu2018and} that exploit temporal information in the learning process are also seen as a separate line of research, with a similar division of methods with respect to the above-mentioned properties. Other aspects in relation to which ReID solutions are considered include different types of challenges where standard methods exhibit certain weaknesses, leading to classes of special, specific methods that effectively overcome the limitations of standard solutions. Thus, the main characteristics of effective ReID solutions aare invariance to different variations of the input signal and robust hierarchical feature representations that lead to accurate identity matching.

Resolution mismatch \cite{jiao2018deep}, when a low resolution query image is matched against the high resolution gallery samples, is a common example of ReID problem that is affected by the scale of the object and image acquisition setup. Opposite situations of high resolution query images are also common, giving rise to resolution independent features \cite{zhang2022resolution} and multiscale pyramid representations \cite{zang2022multidirection}.

The most common person ReID goal is a short-term identification, where it is assumed that the pedestrian's clothing remains unchanged over time. While such assumptions are fully justified in the case of described retail application scenarios, a more general solution would be the invariance of ReID to changes in person's clothing \cite{xu2023deepchange} and appearance \cite{zhang2022illumination}, when captured during different activities. Thus, a step forward ideal long-term or lifelong person ReID would be invariance to cloth-changing, which could be regarded as more applicable in real-world scenarios. In \cite{liu2023learning} authors claim that the most reliable discriminative characteristic for above-mentioned invariance is unclothed body shape, i.e. effective representation of the 3D shape and texture of the human body. It is closely related to other biometric identification methods based on human gait recognition \cite{kwon2021comparative, sepas2022deep}, which are complementary to classical image- or video-based ReID \cite{li2023depth}.

Significant class of adaptation methods also includes occluded person ReID \cite{tan2022dynamic, dong2024multi, tan2024occluded, ning2024enhancement}. A common strategy is to steer ReID toward decisions based on features that comprehend or do not take into account the hidden part of the person. Such examples include learning the dynamic prototype mask \cite{tan2022dynamic}, feature attention mask \cite{DING202091}, saliency-guided patch transfer \cite{tan2024occluded}, using multiple views of the same person \cite{dong2024multi}, or relying on part based representations \cite{agarwal2004learning} and fine-grained image analysis \cite{wei2021fine}, see e.g. \cite{somers2023body, nguyen2021graph}. There have also been attempts to partition feature space \cite{zhai2019defense}, while the standard learning strategy for person ReID has been the triplet loss framework \cite{ding2015deep, hermans2017defense} and its extensions \cite{ijjina2024person}. Alternatives include various deep metric learning approaches \cite{zou2021person} and ranking optimization at inference stage \cite{ ye2021deep}

\section{Proposed person ReID implementation}
\label{Proposed person ReID implementation}

\begin{figure}[htb]
  \centering
  \begin{subfigure}{0.88\columnwidth}
    \centering
    \includegraphics[width=\linewidth]{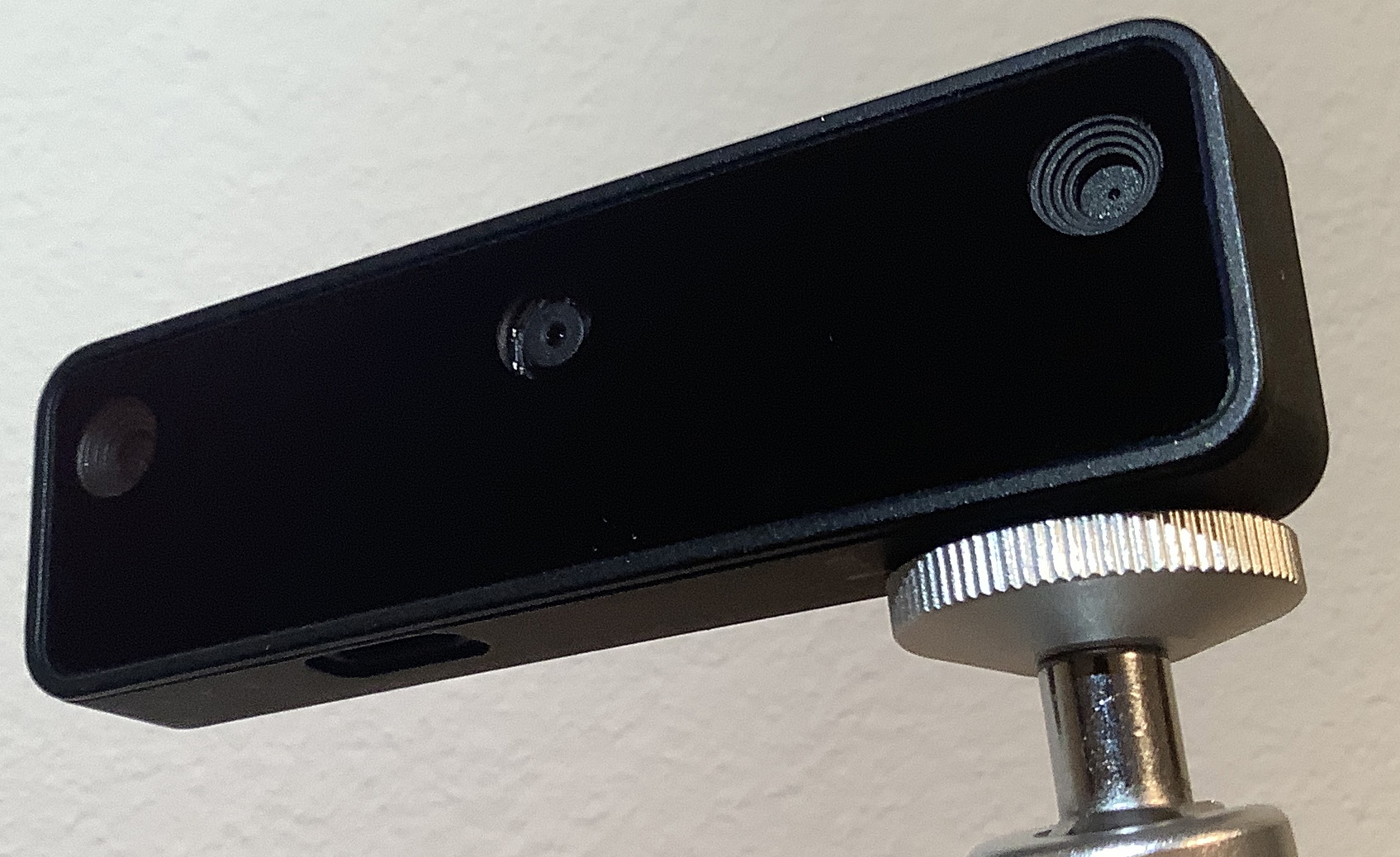}
    \caption{}
    \label{fig:person ReID platform_a}
  \end{subfigure}
  \vspace{1em}

  \begin{subfigure}{0.98\columnwidth}
    \centering
    \includegraphics[width=\linewidth]{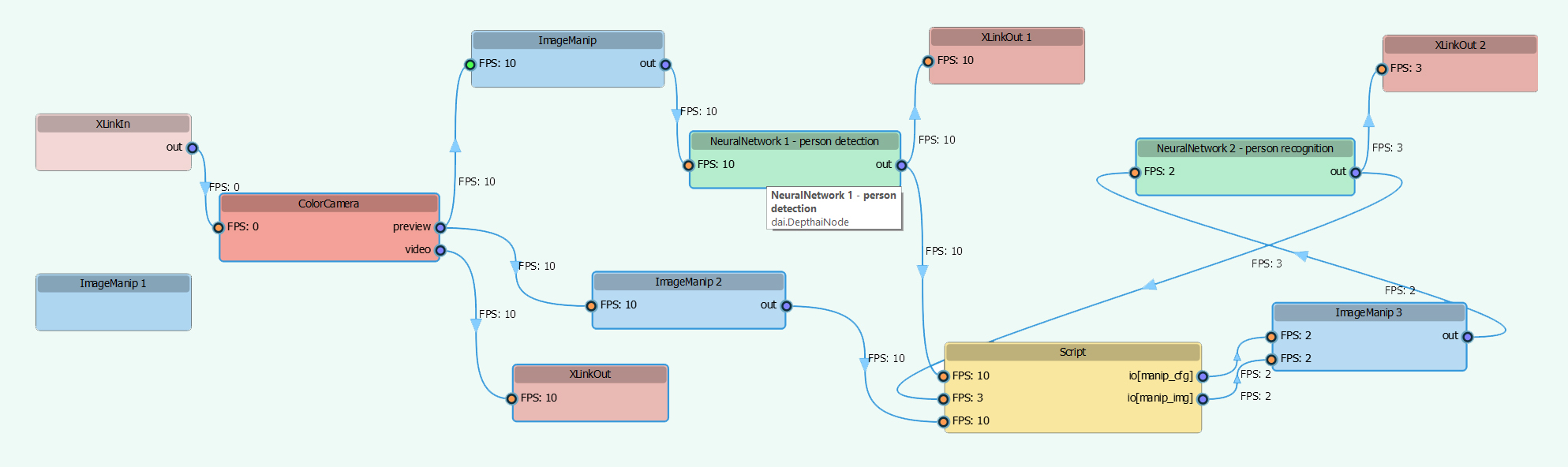}
    \caption{}
    \label{fig:person ReID platform_b}
  \end{subfigure}
  \captionsetup{justification=justified}
  \caption{OAK-D lite \cite{OakDlite2022} vision platform: (a)~three cameras, ISP and AI hardware accelerator on a single board, (b)~implemented processing graph.}
  \label{fig:person ReID platform}
\end{figure}

In order to analyze the sensitivity of person re-identification task under open-world setting of retail stores and public spaces we have implemented one cost efficient, close to real-time solution. It is based on OAK-D lite \cite{OakDlite2022} device and pre-trained person detection and person ReID models made using OpenVINO \cite{openVINO2025} hardware acceleration deep learning framework. The corresponding camera device is shown in Fig.~\ref{fig:person ReID platform_a} and its main characteristics are summarized in Table~\ref{tab:platform_characteristics}.

 \vspace{1em}
\begin{table}[h!]
\caption{Main characteristics of OAK-D lite platform}
\label{tab:oakdlite}
\centering
\begin{tabular}{|p{3cm}|p{5cm}|}
\hline
\textbf{Feature} & \textbf{Specification} \\
\hline
Processor & Intel Movidius Myriad X VPU \\
\hline
Cameras &
\begin{tabular}[t]{@{}l@{}}
1x 13 MP RGB (IMX214, rolling shutter) \\
2x 0.31 MP mono (OV7251, global shutter)
\end{tabular} \\
\hline
Stereo baseline & 75 mm \\
\hline
AI performance & 4 TOPS (1.4 TOPS for AI) \\
\hline
Video output & Up to 4K@30fps \\
 & (H.264/H.265/MJPEG) \\
\hline
Depth perception & Stereo, 300,000-point, 200+ FPS \\
\hline
Connectivity & USB 3.1 Gen1 Type-C \\
\hline
Dimensions (WxHxD) & 91 mm x 28 mm x 17.5 mm \\
\hline
Weight & 61 g \\
\hline
Variants & Fixed-focus / Auto-focus (RGB cam) \\
\hline
Mounting & 1/4"-20 tripod, VESA (7.5 cm, M4) \\
\hline
\end{tabular}
\label{tab:platform_characteristics}
\end{table}
 \vspace{1em}

As can be seen from Table~\ref{tab:platform_characteristics}, it is a versatile and heterogeneous computing platform. It is suitable for applications in different vision tasks that require real time stream processing and advantages of using embedded hardware implementation, such as high level of integration, small form factor, modular design, low power consumption and relatively low cost, taking into account that the device integrates both imaging devices and signal processor. Besides providing the options of choosing between the global and rolling shutter, on board cameras are also designed to be geometrically precisely aligned in a linear arrangement to facilitate calibration and accurate distance measurements. Nevertheless, in the given experiments we have relied only on the main high resolution RGB camera and multiple stream processing by frame grabbing from created camera source in the constructed processing graph, Fig.~\ref{fig:person ReID platform_b}.

 \vspace{1em}

OAK-D lite platform integrates multiple specialized processing units and sensors that collaboratively handle different computational tasks, optimizing performance and efficiency. Thus, Movidius Myriad X Vision Processing Unit (VPU) integrates several system components two of which are of main interest for implemented person ReID. First one is Image Signal Processor (ISP), on-chip hardware accelerator responsible for different camera pipeline functions and high-throughput image pre-processing tasks such as denoising and color correction. The second one is Neural Compute Engine (NCE), neural network hardware accelerator, which enables efficient deployment of trained deep learning models or any other computational graphs that are possible to define in supported ONNX \cite{onnx2025} or OpenVINO \cite{openVINO2025} network exchange formats. Its advantage is small power consumption, which is especially suitable for edge processing and deployment in environments that require continuous system operation. Note that although the main processing (person detection and person ReID) was done on embedded platform, experimental setup also required a host device (CPU) for higher-level program control and application logic (visualization, communication and I/O interface).

For software implementation, besides the device drivers and other utilities provided by Luxonis corporation under the MIT license in official DepthAI code repository \cite{DepthAI2025}, we have used OpenCV \cite{opencv2025} and Python programming language. We note that there are also C++ device interfaces and code implementations, available in \cite{DepthAIc++2025}. The main algorithmic solution was taken from OpenVINO repository and consists of the pre-trained person detection model, available in \cite{openDetectionModel2022}, and person ReID model available from \cite{openReIDModel2022}. As a starting point for developing additional functionalities we have utilized demonstration script in \cite{depthAIReID2023}, from which the provided code implementation was made, Section~\ref{Introduction}. Possible alternative for baseline implementation could also be some general purpose ReID toolbox, like the one proposed in \cite{he2023fastreid}.

\section{Experimental results and discussion}
\label{Experimental results and discussion}

\begin{figure}[!b]
\centering
  \begin{subfigure}{0.31\columnwidth}
    \centering
    \includegraphics[width=\linewidth]{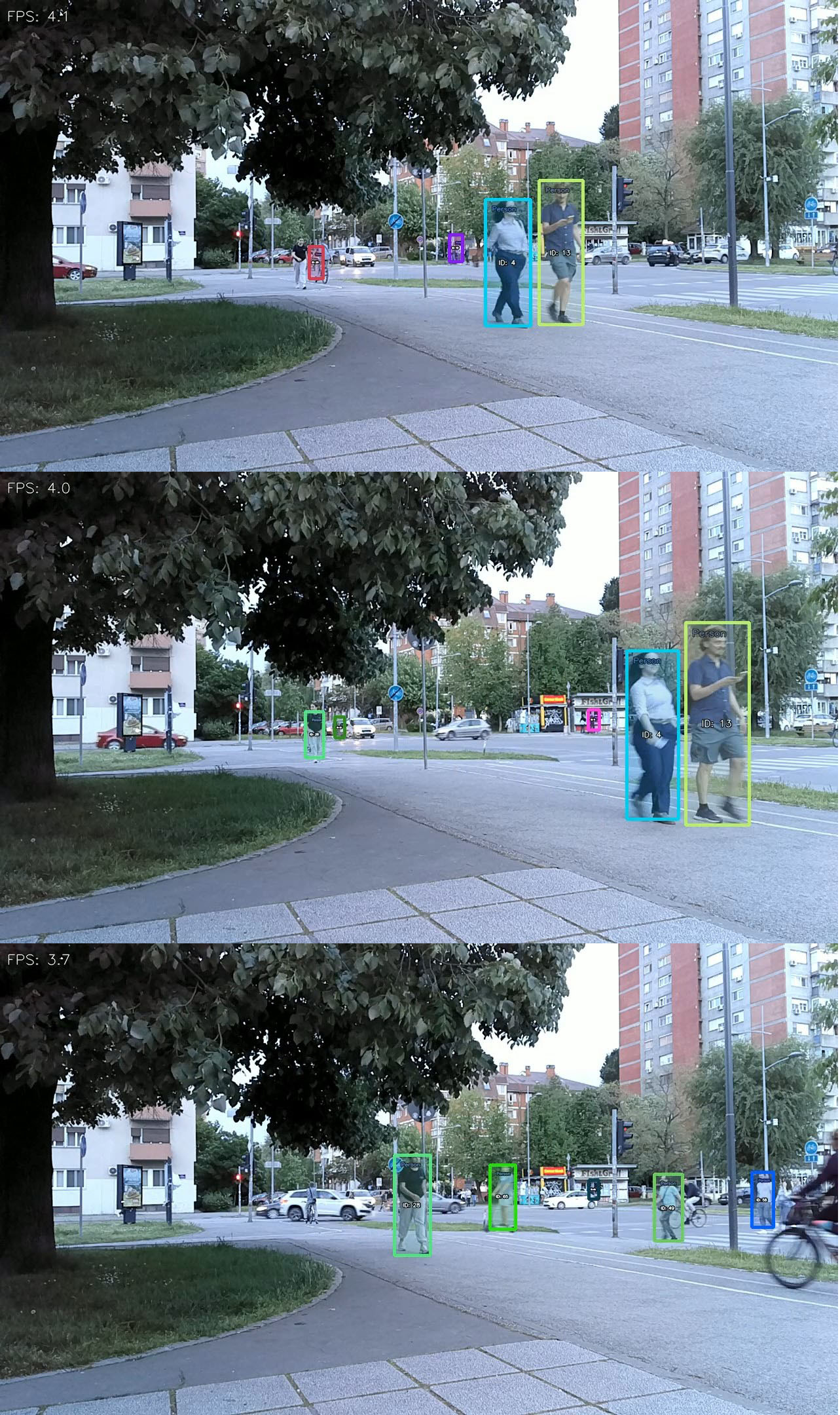}
    \caption{}
    \label{fig:fig2a}
  \end{subfigure}\hfill
  \begin{subfigure}{0.31\columnwidth}
    \centering
    \includegraphics[width=\linewidth]{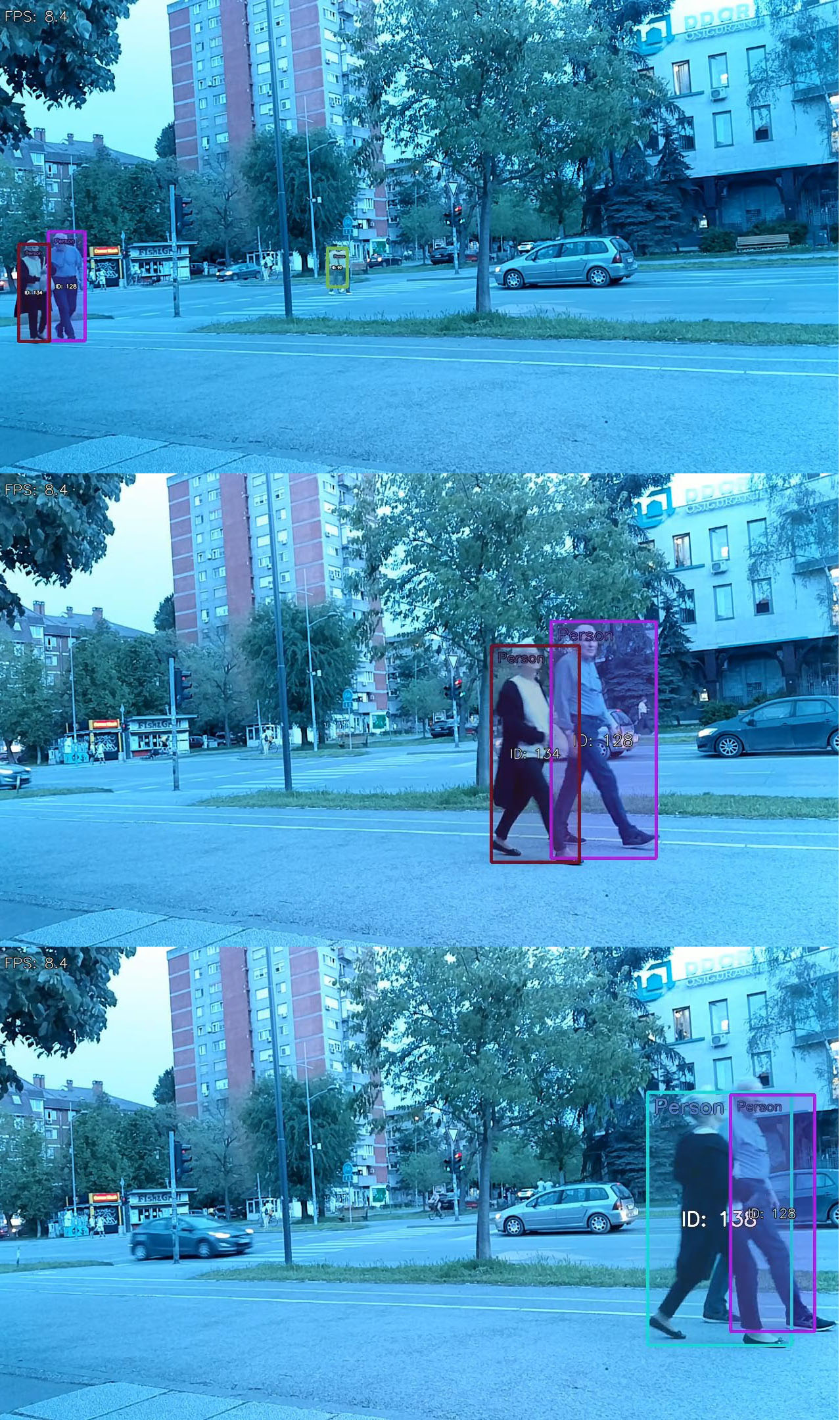}
    \caption{}
    \label{fig:fig2b}
  \end{subfigure}\hfill
  \begin{subfigure}{0.31\columnwidth}
    \centering
    \includegraphics[width=\linewidth]{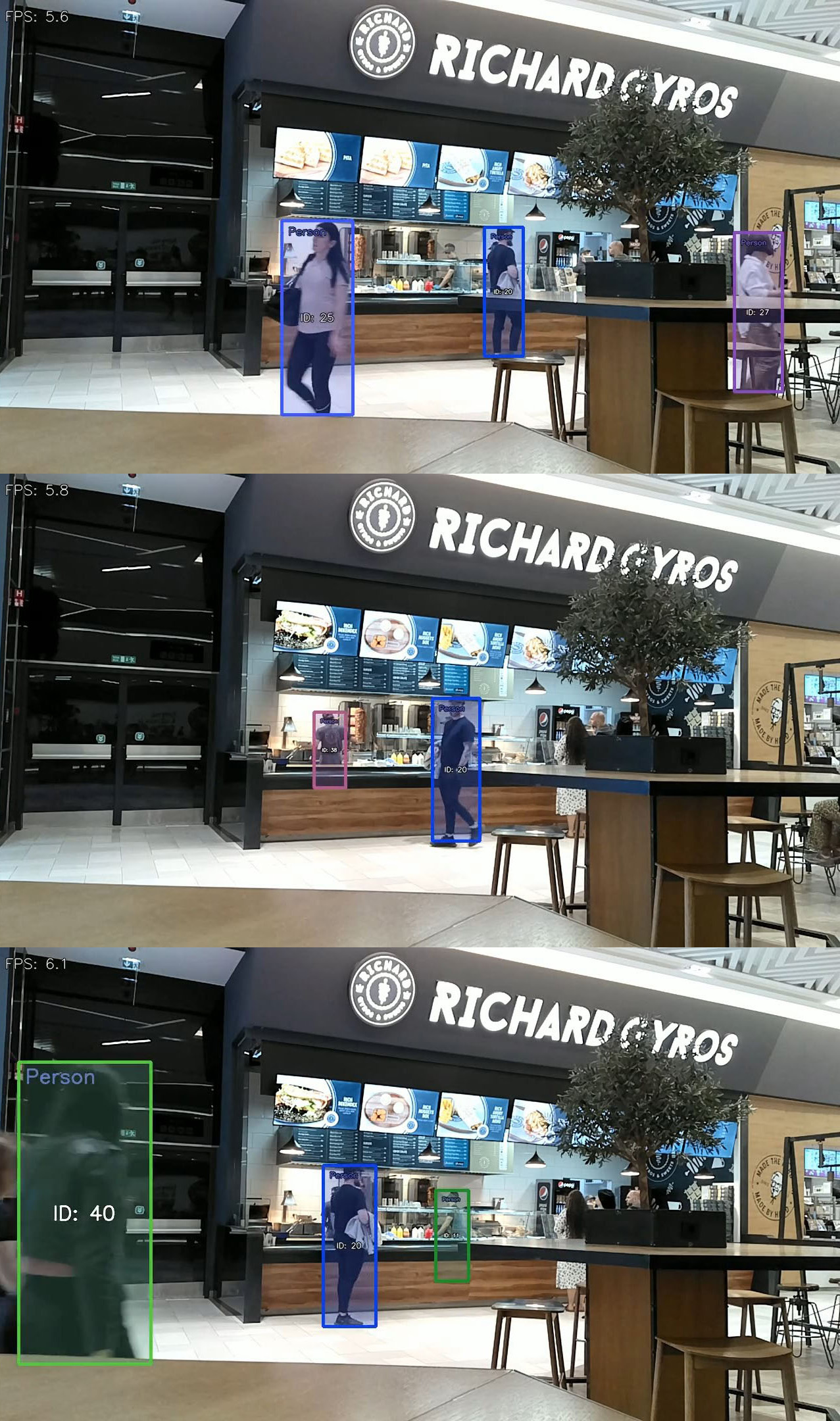}
    \caption{}
    \label{fig:fig2c}
  \end{subfigure}
  \captionsetup{justification=justified}
  \vspace{0em}
  \caption{(a)~succesful ReID, but with low fps rate due to dynamic background noise; (b)~succesful ReID under low light conditions, but with identity loss after change of person's orientation at the end of sequence; (c)~retail store application.}
  \label{fig:experiments1}
\end{figure}

The experiments included visual analysis of system performance in various operating environments. We have analyzed ability of system to cope with person ReID task in outdoor and indoor space, under different illumination and in the presence of varying number and dynamics of people. General conclusions is that in all considered scenarios system managed to achieve close to real time operation, but never more than $\approx 12$ frames per second (fps). On the other hand, under certain difficult conditions, the fps dropped even to $\approx 4$ fps. This included low-light conditions, but mainly the presence of very dynamic background noise in the scene. For example, in an outdoor scene where a significant portion of the background was covered by leaves, in the presence of wind, the background dynamics were very pronounced and caused additional slowdown, Fig.~\ref{fig:fig2a}. However, even in such cases the final ReID result was good and without any false detections in the described background area.

The most challenging ReID were cases where the same person changed orientation towards the camera while walking or when entering and exiting the scene. This led to the creation of multiple identities for the same person, which is a common problem with person ReID. During experiments in an indoor retail space, it was found that certain object arrangements cause false person detection and initialize a correct but false person ReID in subsequent frames. This indicates that person detection module could be subject to further improvements in terms of robustness.

\begin{figure}[!tb]
\centering
  \begin{subfigure}{0.3\columnwidth}
    \centering
    \includegraphics[width=\linewidth]{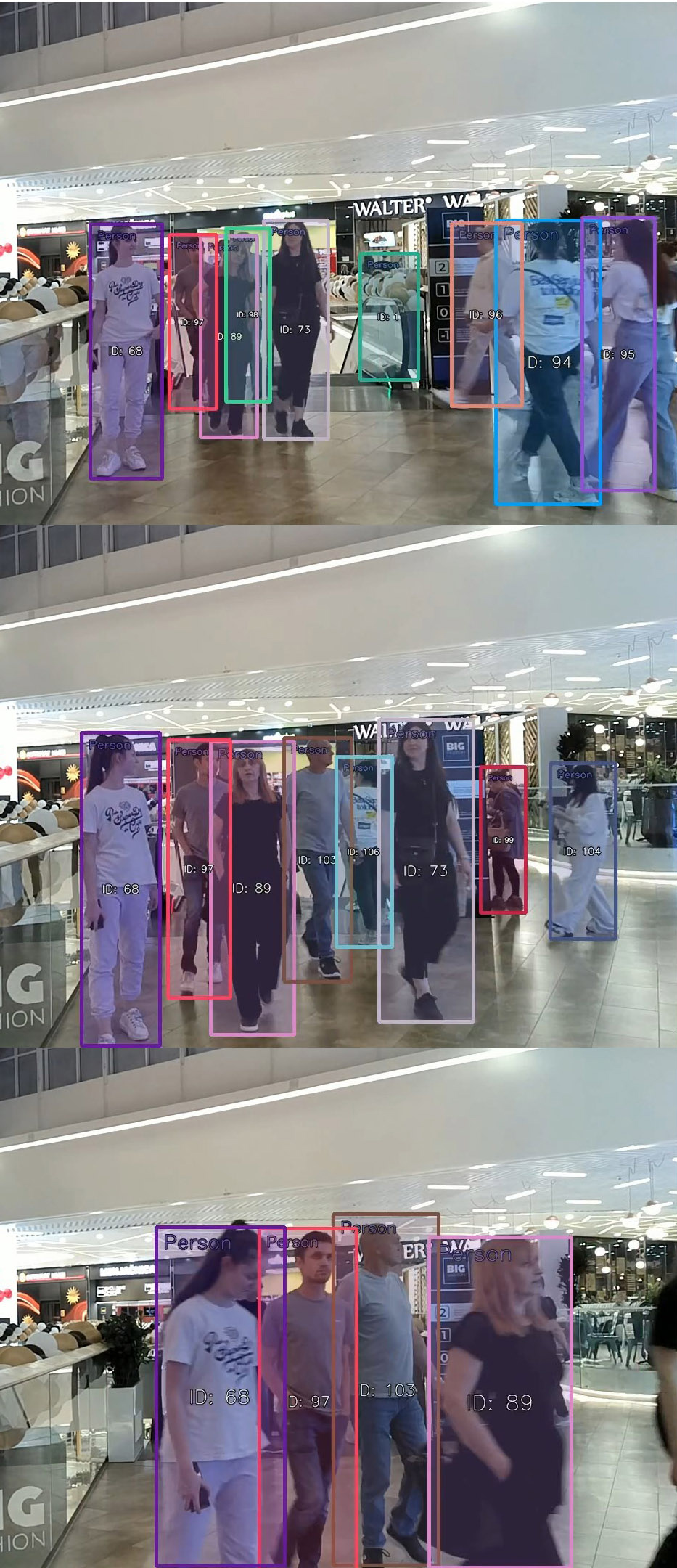}
    \caption{}
    \label{fig:fig3a}
  \end{subfigure}\hfill
  \begin{subfigure}{0.3\columnwidth}
    \centering
    \includegraphics[width=\linewidth]{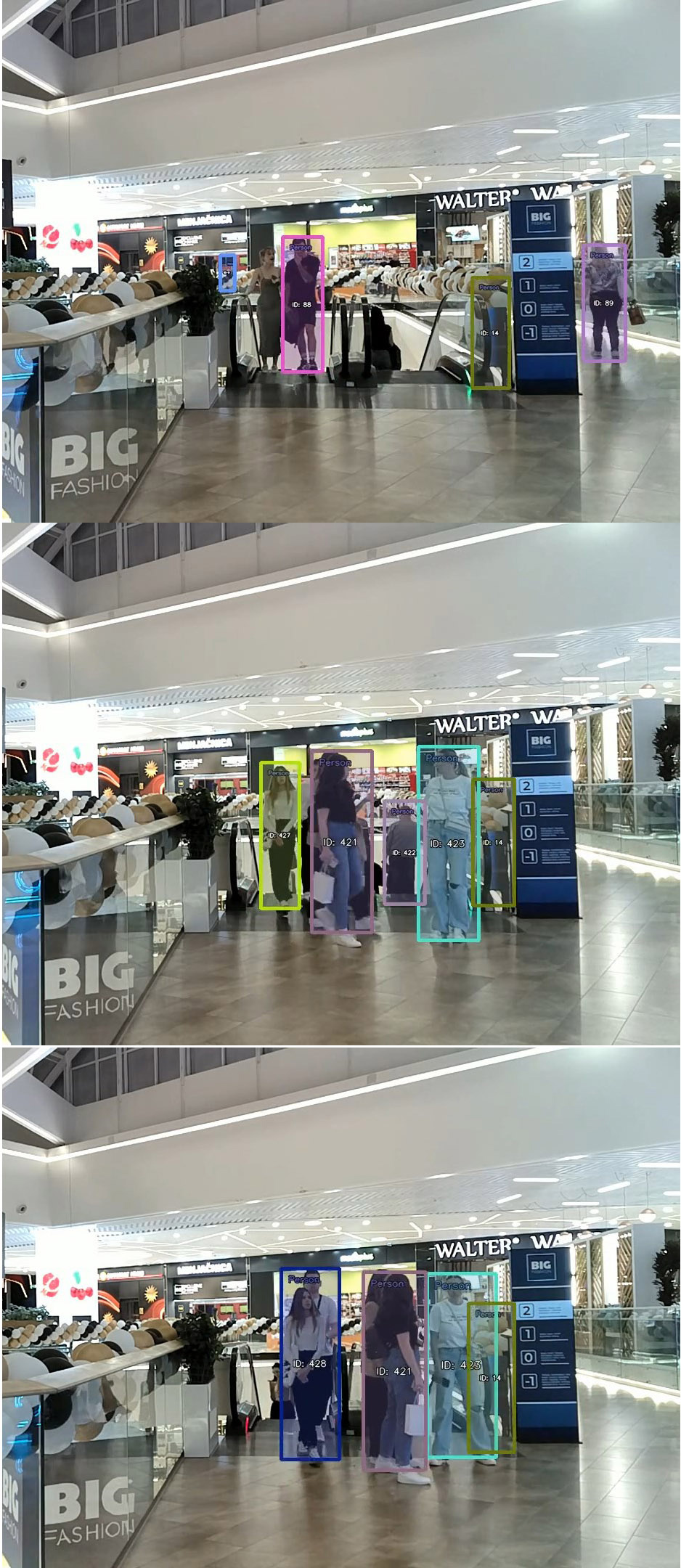}
    \caption{}
    \label{fig:fig3b}
  \end{subfigure}\hfill
  \begin{subfigure}{0.38\columnwidth}
    \centering
    \includegraphics[width=\linewidth]{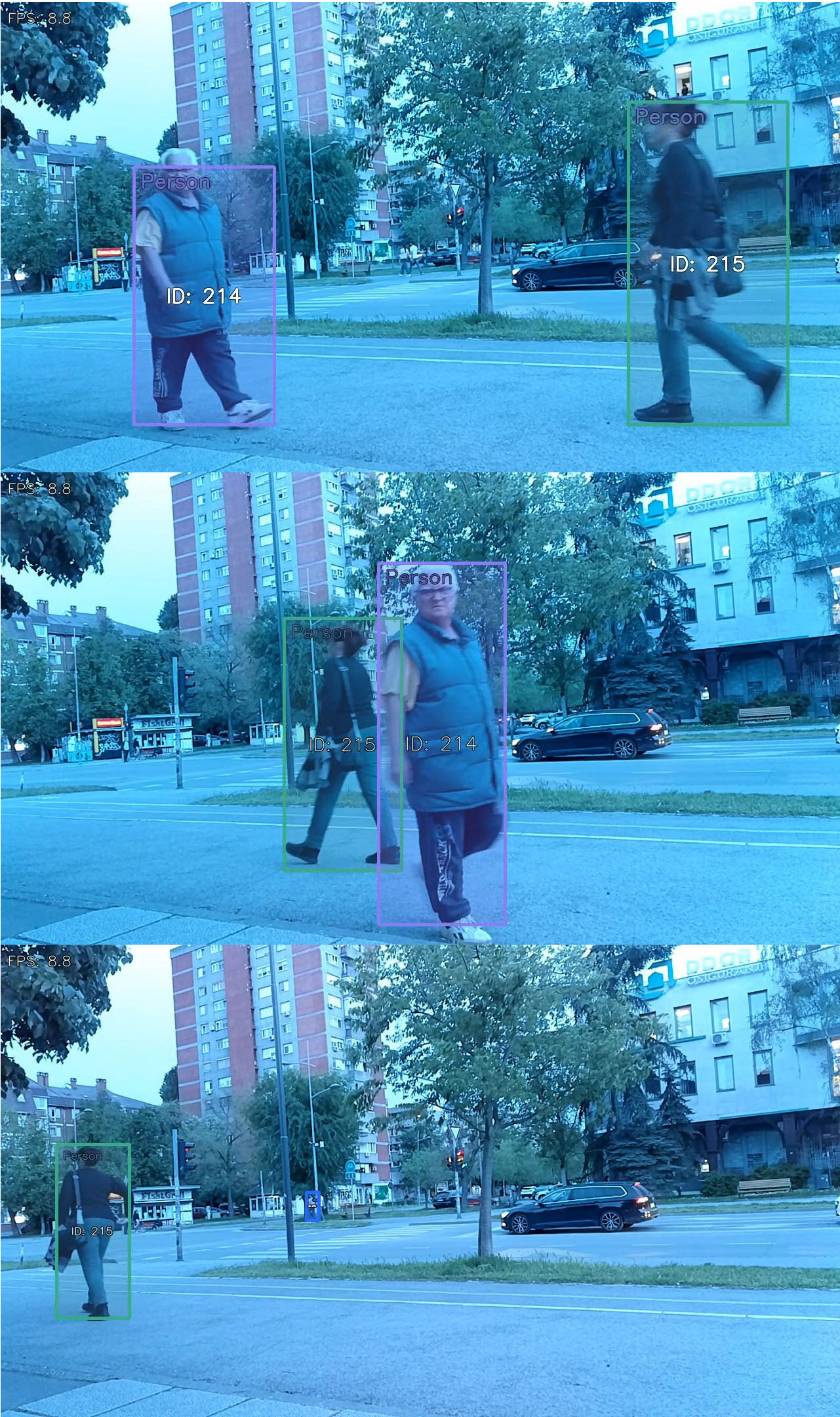}
    \caption{}
    \label{fig:fig3c}
  \end{subfigure}
  \captionsetup{justification=justified}
  \vspace{0em}
  \caption{Experimental results visualization: (a)~crowded indoor scene: successful ReID, but with low fps; (b)~false person detection, but with correct ReID; (c)~low light operation.}
  \label{fig:experiments2}
\end{figure}

Most of the situations described above are illustrated in Fig.~\ref{fig:experiments1} and Fig.~\ref{fig:experiments2}, while the corresponding video files are available in the given code repository, Section~\ref{Introduction}.

The observed variability in people's appearance could be overcome by model training on datasets like \cite{yildiz2024entire, zheng2016mars}, which contain a sufficiently diverse range of appearances (positions of the person relative to the camera, changes in illumination and different human activities). Another research direction are more frequent ReID model updates through better use of historical appereances and without re-indexing of images in the gallery \cite{cui2024learning}.

As already mentioned, crowded scenes and cluttered object detection sometimes require alternative formulations of standard vision tasks \cite{chen2015person, cai2021rethinking}. On the side of possible applications, further analysis of the data collected by ReID systems could reveal a more complex behavior patterns \cite{chophuk2023theft, abbattista2023biometric}, i.e.  go beyond simple statistics like time spent in different parts of a retail store. On the other hand, unsupervised representation learning \cite{chen2022learning} seems to be key to addressing the described challenges and increasing the robustness of ReID.

\section{Conclusions}
\label{Conclusion}
In this paper, we analyzed person re-identification (ReID) task from both technical and application perspectives. Conducted experiments highlighted the need to improve existing methods  in order to make them fully applicable as analytical tools in retail stores and public spaces.  Consequently, incorporating perceptual features such as image depth, audio signals, or other sensing modalities should be expected to enhance ReID performance and reduce false positives in indoor environments. Nevertheless, the general conclusion is that the cost efficient person ReID solutions, like the one analyzed in the paper, are already potentially applicable in the given context.

\section*{Acknowledgment}
The authors would like to thank conference organizers for their support throughout the manuscript preparation process. This research was partially supported by the Science Fund of the Republic of Serbia, project "Multimodal multilingual human-machine speech communication - AI SPEAK", grant no. 7449.

\bibliographystyle{IEEEtranDOI}
\bibliography{ATECH2025}

\end{document}